\documentclass[letterpaper, 10 pt, conference]{ieeeconf}  
\IEEEoverridecommandlockouts                          
\author{Rowan McAllister$^{1}$, Blake Wulfe$^{1}$, Jean Mercat$^{1}$, Logan Ellis$^{1}$, Sergey Levine$^{2}$, Adrien Gaidon$^{1}$
\thanks{$^{1}$Toyota Research Institute
        {\tt\small \{first.lastname\}@tri.global}}%
\thanks{$^{2}$University of California, Berkeley
        {\tt\small svlevine@berkeley.edu }}%
}

\usepackage{biblatex} 
\usepackage{multicol}
\usepackage[makeroom]{cancel}

\usepackage{algorithm}
\usepackage[noend]{algorithmic}
\usepackage{amsmath, amsfonts, amssymb, bbm}    
\usepackage{color, colortbl}
\usepackage[dvipsnames]{xcolor}
\usepackage{hyperref}
\usepackage[capitalize]{cleveref}
\usepackage{caption}
\usepackage{subcaption}
\usepackage{dsfont}

\usepackage{multicol}
\usepackage{multirow}
\usepackage{pifont}  
\definecolor{Gray}{gray}{0.95}

\usepackage{dcolumn}

\usepackage{tikz}
\usetikzlibrary{arrows,shapes,backgrounds,patterns,fadings,decorations.pathreplacing,decorations.pathmorphing, calc}
\tikzset{>=stealth'}

\newcommand{\past}[0]{\mathbf{x}}

\newcommand{\pastspace}[0]{\mathcal{X}}
 
\newcommand{\future}[0]{\mathbf{y}}
\newcommand{\futureego}[0]{\future_{\text{ego}}}
\newcommand{\futureagent}[0]{\future_{\text{agent}}}
\newcommand{\thetaego}[0]{\theta_{\text{ego}}}
\newcommand{\thetaagent}[0]{\theta_{\text{agent}}}
\newcommand{\Future}[0]{\mathbf{Y}}
\newcommand{\futurespace}[0]{\mathcal{Y}}
\newcommand{\probfuturespace}[0]{\mathcal{P}_\futurespace} 
\newcommand{\probfuturespaceego}[0]{\mathcal{P}_{\futurespace_{\text{ego}}}}
\newcommand{\probfuturespaceall}[0]{\mathcal{P}_{\futurespace \times \futurespace_{\text{ego}}}}

\newcommand{\action}[0]{\mathbf{u}}

\newcommand{\actionspace}[0]{\mathcal{U}}

\newcommand{\model}[0]{q_\theta}
\newcommand{\planner}[0]{\pi}
\newcommand{\data}[0]{\mathcal{D}}
\DeclareMathOperator*{\argmax}{arg\,max}

\newcommand{\predfuture}[0]{\hat{\future}}
\newcommand{\predfutureego}[0]{\predfuture_{\text{ego}}}
\newcommand{\predfutureagent}[0]{\predfuture_{\text{agent}}}
\newcommand{\predfuturek}[0]{\hat{\future}^{k}}
\newcommand{\predfuturen}[0]{\hat{\future}_{n}}
\newcommand{\futuren}[0]{\future_{n}}

\newcommand{\predactionk}[0]{\hat{\action}^{k}}

\newcommand{\futureahead}[0]{\future^{\text{ahead}}}
\newcommand{\futurebehind}[0]{\future^{\text{behind}}}
\usepackage{mathtools}
\newcommand{\utility}[0]{\text{utility}}
\newcommand{\attention}[0]{\alpha}
\definecolor{Gray}{gray}{0.9}
 
\newcommand{\citeauthornum}[1]{\citeauthor*{#1}~\cite{#1}}

\title{\LARGE \bf Control-Aware Prediction Objectives for Autonomous Driving}

\begin{document}
\maketitle
\thispagestyle{empty}
\pagestyle{empty}

\begin{abstract}
Autonomous vehicle software is typically structured as a modular pipeline of individual components (e.g., perception, prediction, and planning) to help separate concerns into interpretable sub-tasks. Even when end-to-end training is possible, each module has its own set of objectives used for safety assurance, sample efficiency, regularization, or interpretability. However, intermediate objectives do not always align with overall system performance. For example, optimizing the likelihood of a trajectory prediction module might focus more on easy-to-predict agents than safety-critical or rare behaviors (e.g., jaywalking). 
In this paper, we present 
control-aware prediction objectives (CAPOs), 
to evaluate the \textit{downstream effect} of predictions on control
without requiring the planner be differentiable.
We propose two types of importance weights that weight the predictive likelihood: one using an attention model between agents, and another based on control variation when exchanging predicted trajectories for ground truth trajectories.
Experimentally, we show our objectives improve overall system performance in suburban driving scenarios using the CARLA simulator.
\end{abstract}

\section{Introduction}

Autonomous vehicles (AVs) must navigate busy roads using predictive models to anticipate what surrounding pedestrians and vehicles might do in order to plan safe trajectories around them.
Safe operation requires such components be well calibrated, typically by minimizing some regression error on training data.
However, not all errors made by prediction modules 
are equally important: 
some errors have minimal effect on downstream decisions, while some perceptual errors \cite{teslaCrashReport} and predictive errors \cite{uberCrashReport} can have fatal outcomes.
As no model is perfect, 
it is crucial to identify \textit{which} prediction errors are safety-critical to ensure safety \cite{mcallister2017concrete}.

Whether trained independently or as part of multi-task end-to-end architectures~\cite{P3}, multi-agent trajectory forecasting models typically optimize prediction-specific objectives based on regressing recorded future trajectories by considering all agents equally important \emph{a priori}.
However, when considering the target control task of autonomous navigation, some predictions warrant more attention than others when deciding safe controls. Consequently, control-agnostic optimizing of prediction models may not result in improved downstream navigation performance due to limited data, model capacity, rare events, or computational constraints.
Even with end-to-end training, multi-task objectives might not be aligned, thus resulting in performance degradation due to task interference~\cite{wu2020understanding}.

\begin{figure}[t]
    \centering
    \includegraphics[width=\linewidth]{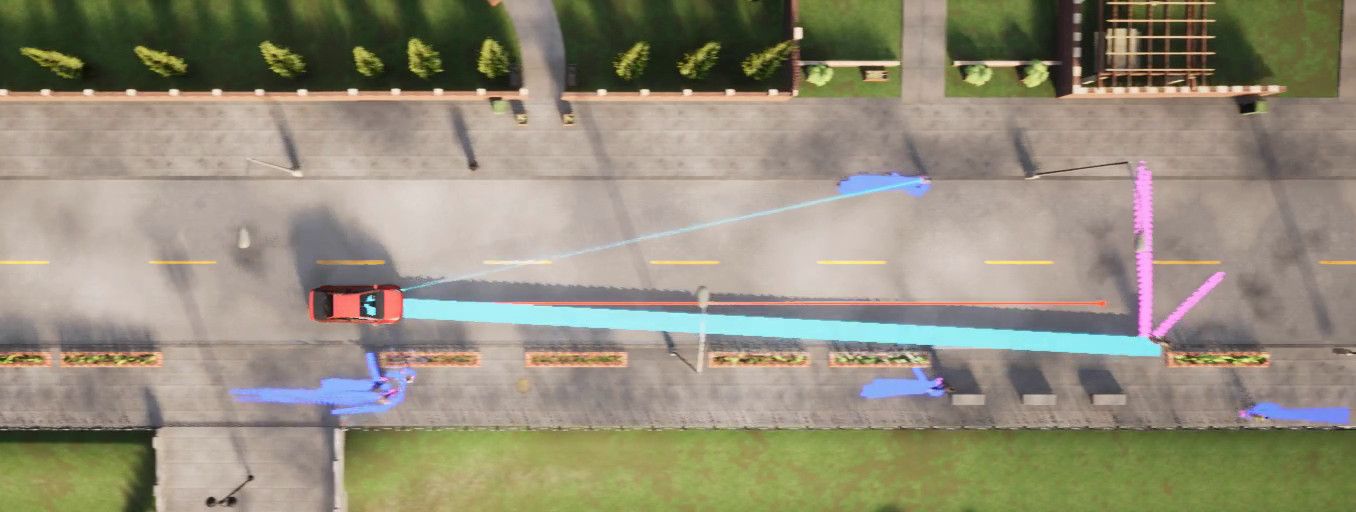}
    \caption{A vehicle drives to the right while reacting to pedestrians with sample predicted trajectories shown in purple or pink. Our Control-Aware Prediction Objectives (CAPO) can learn to capture which predictions should have more influence on the vehicle's controls (cyan lines proportional to attention).
    Videos available at \url{https://sites.google.com/view/control-aware-prediction}
    }
    \label{fig:teaser}
    \vspace{-10pt}
\end{figure}

In this work, we propose \emph{Control-Aware Prediction Objectives} (CAPOs) to train prediction models that more accurately reflect the relative effects of predictive errors on downstream control. Computing these downstream effects requires only forward passes without backpropagation between modules. This improves applicability with real-world AV planning and control systems,
which might not be fully differentiable due to complex design constraints (e.g., verifiability, interpretability, comfort and safety constraints).
Our method introduces importance-weighted prediction likelihood objectives using forward passes of the prediction model and planner. We investigate two weighting methods that can be trained with backpropagation. The first assigns weights based on control variations due to prediction changes.
The second uses learned attention weights between agent predictions and AV controls.

Using the CARLA simulator, we experimentally show that training prediction models with control-aware objectives leads to improved controller performance in complex multi-agent urban driving scenarios. Compared with existing prediction models, including prediction algorithms that treat everything as equal, we show that our new objective helps to avoid precisely those errors that would maximally influence downstream decisions. 

\section{Related Work} \label{sec:related-work}

Several related fields of study investigate objective-aware prediction metrics as we discuss here.

\subsection{Objective-Aware Prediction in Reinforcement learning}

Model-based reinforcement learning (MBRL) methods learn a dynamics model of an autonomous agent to predict which control decisions lead to states with higher objective rewards \cite{deisenroth2011pilco,nagabandi2018learning,moerland2020model}.
MBRL prediction is related to AV prediction, with the main difference being that AVs predict the trajectories of \textit{other} human agents and not those of the autonomous agent.
Nevertheless, several MBRL works have recently challenged the common assumption that the better a dynamics model's predictive accuracy, the better the downstream policy will maximize reward. For example, \citeauthornum{lambert2020objective} show task-agnostic loss functions used to train dynamics models are often uncorrelated with episode rewards, an issue termed ``objective mismatch''.

Indeed, learned dynamics models need not be accurate everywhere in the state space, only in the areas that help maximize rewards \cite{bansal2017goal}. 
Some RL works investigate training models insofar as they improve estimating the value function \cite{farahmand2017value,ayoub2020model}, policy gradient \cite{joseph2013reinforcement,abachi2020policy},
or ability to reach a goal state \cite{nair2020goal}.
Others optimize downstream policies directly using Bayesian optimization to search model parameters \cite{bansal2017goal}.
Work by \citeauthornum{donti2017task} points out that in practice we often want a combination of training a model to optimize its likelihood as well as a downstream task term, although in the context of constrained optimisation.
Similarly, \citeauthornum{lambert2020objective} correlate both metrics by increasing the weight in the loss of data points closer to data the optimal controller generates.

Unfortunately for AV applications, such objective-aware prediction methods are often inapplicable for several reasons. First, MBRL assumes access to an objective reward function, reward samples, or goal state, but objective measures or goals of desirable driving are often difficult to define. By contrast, human-designed AV control systems are often preferable for verification and interpretability reasons.
Second, a common assumption in RL is that the policy is either differentiable or stochastic (in the case of policy gradients), whereas real-world AV control systems often contain complex logic that is neither differentiable nor stochastic. 
Our work focuses instead on how to learn prediction  
given access to a safe, potentially non-differentiable controller.

\subsection{Map-Aware Prediction Metrics} 

Map information can help incorporate prior knowledge into prediction metric design. For example, since the ego vehicle drives on the road, pedestrian forecasting errors could be given more weight on road surfaces than otherwise. Work by \citeauthornum{shridhar2020beelines} use maps to help focus on potential collisions with other agents by generating a set of candidate ego trajectories along known lane tracks that any controller might follow.
This method does not assume a particular downstream controller, but makes an educated guess as to what a reasonable controller might do.
Another map-based prediction metric is Drivable Area Compliance (DAC) \cite{chang2019argoverse}, which counts the proportion of model samples that exit the drivable area.
A conceptual difference with our method is that our prediction metrics assume access to the specific downstream controller that will be used at test-time, improving test-time performance.
Since controllers already consider road information, we circumvent the need to explicitly design prediction metrics around mapping information, which can incur additional hyperparameters, such as the relative costs of predicting if an agent will traverse either road / sidewalk / building. 

\subsection{Control-Aware Perception}

\citeauthornum{philion2020learning} 
propose a control-aware 3D object perception metric called Planning KL-divergence (PKL) based on how perceptual errors cause distributional divergence in the ego's distribution of planned paths, compared to a planner with perfect observations, measured by the KL divergence. 
In contrast, our work focuses on prediction objectives, and additionally demonstrates how the new objective empirically affects online-control using a driving simulator.
We also avoid KL distance losses in our work since this assumes non-trivial stochasticity in the controller or data, which is not always the case.
Other works have also used KL policy distances to investigate how observations can be compressed while preserving human-like actions had they remained uncompressed \cite{reddy2021pragmatic}.
Work by \citeauthornum{piazzoni2020modeling} also investigates how perceptual metrics affect downstream planning but are specific to perception.

\section{Preliminaries} \label{sec:preliminaries}

Here we formalize our notation and discuss some existing prediction metrics before presenting our own.

\subsection{Notation and Assumptions}

\newcommand{\sample}[0]{\sim}

Let $\past\in\pastspace$ denote past trajectory information about all agents, used to make probabilistic predictions $\predfuture\in\probfuturespace$ about the future multi-agent trajectories $\future\in\futurespace$. Trajectories are predicted up to time horizon $T$, and $\future_T$ denotes the future state at time $T$.
As the intents of other agents are usually uncertain, we use a \textit{probabilistic} prediction model $\model$ with trainable parameters $\theta$ to sample the motion of others: 
$\predfuture\sample\model(\Future|\past)$, denoting likelihoods as $\model(\future|\past)\doteq\model(\Future=\future|\past)$.
If multiple samples are taken, $\predfuturek$ refers to the $k$th sample, 
and to single out the $n$th agent we overload notation using $\predfuturen$, and use $\futureego$ as the AV's future trajectory. Given such predictions, the AV controller $\planner$ outputs ego controls $\action\in\actionspace$ to anticipate and avoid colliding with other agents' future trajectories: $\action = \planner(\future)$.

We assume our AV stack performs behavior prediction before control, a common assumption \cite{schwarting2018planning}. While conditioning behavior prediction on ego's intent provides more accurate prediction, for sake of simplicity we assume that other agents do not anticipate the AV's future, only the AV anticipates the other agents' future trajectories in order to avoid collisions.

\subsection{Common Prediction Metrics and Objectives}
Common prediction metrics in the literature and in prediction benchmarking challenges--including Argoverse Forecasting~\cite{chang2019argoverse}, Lyft Prediction~\cite{houston2020one}, Waymo Open Motion~\cite{ettinger2021large}, and AIODrive~\cite{weng2021all}--are summarized in \cref{tab:metrics}:

\begin{table}[H]
\centering
\caption{Common prediction metrics in the literature.}
\label{tab:metrics}
\begin{tabular}{lc} 
 \hline
 Metric Name & Metric Equation \\
 \hline
 Average Displacement Error (ADE) & $||\predfuture - \future||_2$ \\ 
 Final Displacement Error (FDE) & $||\predfuture_T - \future_T||_2$ \\
 Minimum-ADE (minADE)  & $\min_{k\in[K]}||\predfuturek - \future||_2$ \\ 
 Minimum-FDE (minFDE) & $\min_{k\in[K]}||\predfuturek_T - \future_T||_2$ \\
 Miss Rate (MR) & $\frac{1}{K}\sum_{k}\mathds{1}[||\predfuturek_T - \future_T||_2 > \alpha]$ \\
 Negative Log Likelihood (NLL): & $-\log \model(\future|\past)$ \\ 
 \hline
\end{tabular}
\end{table}

Most metrics compare the Euclidean distance between either the full predicted state-sequence $\predfuture$ (or final state $\predfuture_{T}$) with the true sequence $\future$ (or final state $\future_{T}$) an agent took, as recorded in data.
Probabilistic models are typically trained to minimize the negative log likelihood (NLL) of the data. 
All such metrics are agnostic to road geometry 
and downstream planning,
which implicitly assumes that all other agents' forecasts are equally relevant.
For example, consider two pedestrians: one walking ahead of the ego vehicle and one behind.
Assuming independent pedestrian motion, the NLL objective factorizes as: $-\log \model(\futureahead, \futurebehind|\past) = -\log \model(\futureahead|\past) - \log \model(\futurebehind|\past)$.
Notice that this prediction metric is \textit{equally} concerned with both $\futureahead$ and $\futurebehind$.
Intuitively, accurate prediction of the pedestrian ahead of the ego vehicle is more important for safe motion planning since the ego's planned path is more likely to intersect with $\futureahead$ than $\futurebehind$. 
How can prediction metrics become ``aware'' that errors in predicting $\futureahead$ have greater downstream consequences than errors in $\futurebehind$?

\section{CAPO: Control-Aware Prediction Objectives} \label{sec:method}

In this section we propose novel prediction loss functions that consider \textit{how} predictions will be used downstream to improve predictive accuracy whenever prediction errors would cause a large change in control outputs. 
In Bayesian decision theory, a decision is evaluated as the expected utility of a decision $\action$ or controller $\planner$, integrating out any uncertainties \cite{barber2012bayesian}. In our case, it is the future trajectories of other agents that are unknown but can be probabilistically predicted according to a model with parameters $\theta$. Following the literature on loss-calibrated variational inference \cite{lacoste2011approximate,cobb2018loss,kusmierczyk2019variational}, we define the \textit{gain} of a decision or controller's value as a function of the model parameters $\theta$ that we wish to train.
\begin{align}
\text{Gain}_{\past, \future, \planner}(\theta) \;&=\; \int \utility(\planner,\future,\predfuture,\past)\model(\predfuture|\past)  \text{d}\predfuture. \label{eq:gain}
\end{align}  

The choice of utility function in \cref{eq:gain} is an open one, that is why we considered many possible input parmeter; it defines how desirable a course of actions would be given $\past$ and $\predfuture$. Alternatively, an existing metric like the NLL can simply be weighted without integration. In the next subsection we discuss some baseline choices for the utility or weight, and after, we propose two novel methods for computing these weights: a \textit{self-attention} method and a \textit{counterfactual} method.

\subsection{Baseline Objectives}

Most predictive metrics in the literature are agnostic to $\action$ and simply use a delta function to only score correct trajectory predictions, recovering the standard log likelihood metric:
$\text{Gain}_{\past, \future}(\theta)
= \int \delta(\future-\predfuture)\model(\predfuture|\past)  \text{d}\predfuture
= \model(\future|\past)$. 
However, we are interested in utilities that are a function of $\action$ in order to weight predictions by their downstream effect on the ego's control. 
For instance, we could score trajectory predictions based on the resultant ego controls $\planner(\predfuture)$ matching the ego's behavior under knowledge of the true future trajectories $\planner(\future)$:
$\text{Gain}_{\past, \future, \planner}(\theta) = \int \delta(\planner(\future)-\planner(\predfuture)) \model(\predfuture|\past) \text{d}\predfuture.$ 
This integral is unfortunately intractable to derive or estimate, but softer utility functions can be used instead. One example is $||\planner(\predfuture) - \planner(\future)||_1$, which we include as a baseline in \cref{table:ourmethods}.
Optimizing this controller output error guides the learning process towards predicting controller inputs (predicted trajectories) accurately, insofar as they result in the correct control. Any trajectory errors that do not induce a change in the AV's control are thus considered inconsequential and ignored.

\subsection{Attention-based CAPO} \label{sec:attention}
We propose a GRU encoder-decoder architecture with an attention mechanism as introduced in \cite{vaswani2017attention}. Our method weights the agent predictions using attention factors between agents $\past$ and the AV's future trajectory $\futureego$. The predictive model is a function parameterized by $\theta$ noted $\model:\pastspace\to\probfuturespaceall$. We note $\theta = \{\thetaego, \thetaagent\}$ where $\thetaego$ is the set of parameters for the ego decoder only; $\pastspace$ is the past observation space and $\probfuturespaceall$ 
the probability spaces of future trajectories: $\probfuturespaceego$ for the ego and $\probfuturespace$ for other agents.

We use an architecture similar to \cite{leurent2019social, mercat2020multi} where we train a model with multi-head attention; the ego agent attends the other agents. The ego predictions are used as a proxy for the actual planner to compute the importance weights of other agents:

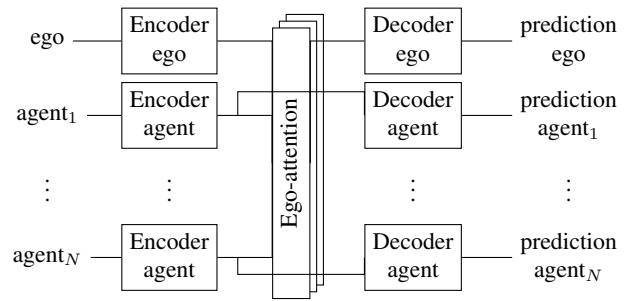
\begin{figure}[H]
	\centering
	\begin{tikzpicture}[scale=0.9, every node/.style={scale=0.9, align=center}]
	\node(X1){ego};
	\node[below of=X1, node distance=1.1cm](X2){agent$_{1}$};
	\node[below of=X2, node distance=1cm](X3){$\vdots$};
	\node[below of=X3, node distance=1.1cm](X4){agent$_{N}$};
	
	\node[draw, right of=X1, node distance=1.8cm, rectangle](ENC1){Encoder \\ ego};
	\node[draw, right of=X2, node distance=1.8cm, rectangle](ENC2){Encoder \\ agent};
	\node[below of=ENC2, node distance=1cm](ENC3){$\vdots$};
	\node[draw, right of=X4, node distance=1.8cm, rectangle](ENC4){Encoder \\ agent};
	
	\path (X1) edge (ENC1);
	\path (X2) edge (ENC2);
	\path (X4) edge (ENC4);

	\node[draw, rectangle, right of=ENC1, node distance=2.0cm, below=-0.4cm, minimum height=4cm, fill=white](TRANS3){\rotatebox{90}{ Ego-attention }};
	\node[draw, rectangle, right of=ENC1, node distance=1.9cm, below=-0.3cm, minimum height=4cm, fill=white](TRANS2){\rotatebox{90}{ Ego-attention }};
	\node[draw, rectangle, right of=ENC1, node distance=1.8cm, below=-0.2cm, minimum height=4cm, fill=white](TRANS1){\rotatebox{90}{ Ego-attention }};
	
	\coordinate[right of=ENC2, node distance=1.9cm, below=-0.35cm](THROUGH2){};
	\coordinate[right of=ENC4, node distance=1.9cm, below=0.25cm](THROUGH4){};

	\draw (ENC1.east) -| (TRANS1.west);
	\draw (ENC2.east) -| (TRANS1.west);
	\draw (ENC4.east) -| (TRANS1.west);

	\draw ($(ENC2.east)+(0.3, 0)$) |- (THROUGH2);
	\draw ($(ENC4.east)+(0.3, 0)$) |- (THROUGH4);
	
	\node[draw, right of=ENC1, node distance=3.6cm, rectangle](DEC1){Decoder \\ ego};
	\node[draw, right of=ENC2, node distance=3.6cm, rectangle](DEC2){Decoder \\ agent};
	\node[below of=DEC2, node distance=1cm](DEC3){$\vdots$};
	\node[draw, right of=ENC4, node distance=3.6cm, rectangle](DEC4){Decoder \\ agent};
	
	\draw (TRANS1.east) |- (DEC1.west);

	\node[right of=DEC1, node distance=2.3cm](Y1){prediction \\ ego};
	\node[right of=DEC2, node distance=2.3cm](Y2){prediction \\ agent$_1$};
	\node[below of=Y2, node distance=1cm](Y3){$\vdots$};
	\node[right of=DEC4, node distance=2.3cm](Y4){prediction \\ agent$_N$};
	
	\draw (THROUGH2) -| (DEC2.west);
	\draw (THROUGH4) -| (DEC4.west);

	\draw (DEC1.east) -- (Y1.west);
	\draw (DEC2.east) -- (Y2.west);
	\draw (DEC4.east) -- (Y4.west);
	\end{tikzpicture}
	\caption{Diagram of the attention model. All agent encoders and decoders share their weights. Encoders and decoders are GRUs. Attention is not used between agents.}
	\label{fig:architecture}
	
\end{figure}

The ego-attention blocks in figure \cref{fig:architecture} are heads of a multi-head attention mechanism. The computation performed by each head is given below:

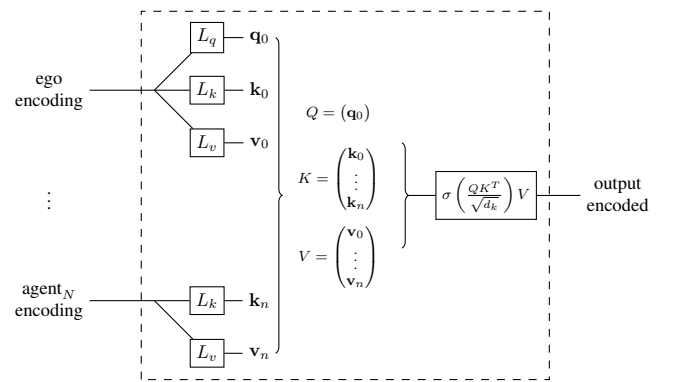
\begin{figure}[H]
	\centering
	\begin{tikzpicture}[scale=0.7, every node/.style={scale=0.7}]
	\node(X1){$\begin{matrix}\text{ego}\\\text{encoding}\end{matrix}$};
	\node[below of=X1, node distance=2cm](X2){$\vdots$};
	\node[below of=X2, node distance=2cm](X3){
	$\begin{matrix}\text{agent}_{N}\\\text{encoding}\end{matrix}$};
	
	\coordinate[right of= X1, node distance=2cm](X1b){};
	
	\draw (X1) -- (X1b);
	
	\node[draw, right of=X1b, node distance=1cm](LK1){$L_{k}$};
	\node[draw, below of=LK1, node distance=1cm](LV1){$L_{v}$};
	\node[draw, above of=LK1, node distance=1cm](LQ1){$L_{q}$};
	
	\draw (X1b) -- (LQ1);
	\draw (X1b) -- (LK1);
	\draw (X1b) -- (LV1);

	\node[right of=LQ1, node distance=1cm](Q1){$\mathbf{q}_0$};
	\node[right of=LK1, node distance=1cm](K1){$\mathbf{k}_0$};
	\node[right of=LV1, node distance=1cm](V1){$\mathbf{v}_0$};
	
	\draw (LQ1) -- (Q1);
	\draw (LK1) -- (K1);
	\draw (LV1) -- (V1);
	
	\coordinate[right of= X3, node distance=2cm](X3b){};
	
	\draw (X3) -- (X3b);
	
	\node[draw, right of=X3b, node distance=1cm](LK3){$L_{k}$};
	\node[draw, below of=LK3, node distance=1cm](LV3){$L_{v}$};

	\draw (X3b) -- (LK3);	
	\draw (X3b) -- (LV3);
	
	\node[right of=LK3, node distance=1cm](K3){$\mathbf{k}_{n}$};
	\node[right of=LV3, node distance=1cm](V3){$\mathbf{v}_{n}$};

	\draw (LK3) -- (K3);
	\draw (LV3) -- (V3);
	
	\coordinate[right of=Q1, node distance=0.3cm](TOP){};
	\coordinate[right of=V3, node distance=0.3cm](BOT){};
	\draw[decorate,decoration={brace}] (TOP) -- node[left=5pt]{} (BOT);
	
	\node[right of=X2, text width=3cm, node distance=5.5cm](EQ){
		\footnotesize \[Q = \left( \begin{matrix}
		\mathbf{q}_0
		\end{matrix} \right)\]
		\\
		\footnotesize \[K = \left( \begin{matrix}
		\mathbf{k}_0 \\
		\vdots \\
		\mathbf{k}_{n}
		\end{matrix} \right)\]
		\\
		\footnotesize \[ V = \left( \begin{matrix}
		\mathbf{v}_0 \\
		\vdots \\
		\mathbf{v}_{n}
		\end{matrix}\right) \]
	};

	\node[draw, right of=X2, node distance=8.3cm](EQ2){
	\footnotesize $\sigma\left(\frac{QK^T}{\sqrt{d_k}}\right)V$};

	\coordinate[left of=EQ2, node distance=1.6cm, above=1.0cm](TOP2){};
	\coordinate[left of=EQ2, node distance=1.5cm, below=0.0cm](MID2){};
	\coordinate[left of=EQ2, node distance=1.6cm, below=1.0cm](BOT2){};
	\draw[decorate,decoration={brace}] (TOP2) -- node[left=5pt]{} (BOT2);
	\draw (MID2) -- (EQ2);
	
	\node[right of=EQ2, node distance=2.5cm](OUT){$\begin{matrix}\text{output}\\\text{encoded}\end{matrix}$};
	\draw (EQ2) -- (OUT);
	
	\draw[draw=black, dashed] (1.75cm, -5.5cm) rectangle (9.5cm,1.5cm);
	\end{tikzpicture}
	
	\caption{Diagram of a self-attention head.}
	\label{fig:head}
\end{figure}

The attention vector is given by 
\begin{equation}
\attention = \sigma\left(\frac{QK^\top}{\sqrt{d_k}}\right) = [\attention_0,...,\attention_N] \label{eq:attention},
\end{equation}
where $Q$ is the query matrix, $K$ the key matrix, and $\sigma$ the softmax operation that normalizes the attention vector (for details see \cite{vaswani2017attention}). The encoded $\text{output} = \attention V$ is a weighted mean of the value vectors over the agents (including ego).

Inspired by \cite{mercat2020multi} the attention model produces outputs in the form of a sequence of Gaussian mixtures for each agent and is trained to minimize the NLL for all agents and the ego trajectory predictions. However, for our application, ego prediction is not the goal, it is only a proxy to compute the importance weights of the pedestrian contribution to the ego behavior.

We propose to use attention coefficients $\attention$ as importance factors in a weighted sum of per-human state prediction loss (as opposed to uniform weighting). Algorithm \ref{algo:ourattentionmethod} summarizes how the model is trained with importance weighting. If multiple heads are used, the attention coefficients are averaged:
\begin{align}
w_n = \frac{1}{H}\sum_{h=1}^H\attention_n^{(h)}
\label{eq:weight_attention}
\end{align}
The attention predictor imitates a planner that interacts with the other agents to avoid collision.
The only way for the predictor to interact with the other agents is through attention.
Therefore, as the model learns the correlations between the planner's trajectories and the agents trajectories, larger attention coefficients are given to the agents that cause larger reactions from the controller.
It learns this offline and does not need access to the controller nor its gradient.
 
Predicting jointly the ego trajectory and the other agents allows us to use the attention coefficients for concern weighting in a single run. The coefficient $\attention_0$ quantifies how independent the ego is from other agents.

This method defines a concern about an agent but not about specific trajectories of that agent. It can define the concern without using the controller because it instead uses an offline-learned model that imitates the controller.
\begin{algorithm}[H]
    \textbf{Input}: Controller: $\planner:\pastspace\to\actionspace$
    \begin{algorithmic}[1]
     \STATE{Record trajectory data $\data=\{\past,\future\}_i$}
     \WHILE{training}
        \STATE{Sample batch $\past,\future\sample\data$}
        \STATE{Run model to estimate $\predfutureego$ and $\predfutureagent$ from $\past$}
        \STATE{Get attention: $\attention(\past)$ \hfill $\triangleright$ \cref{eq:attention}}
        \STATE{Compute weight: $w(\attention(\past))$ \hfill $\triangleright$ \cref{eq:weight_attention}}
        \STATE{Update model: \\
        $\thetaego\leftarrow\thetaego+\nabla_{\thetaego}\log\model(\futureego|\past)$\\
        $\thetaagent\leftarrow\thetaagent+w(\past)\nabla_{\thetaagent}\log\model(\futureagent|\past) 
        $}\\
     \ENDWHILE
    \end{algorithmic}
    \textbf{Output}: Predictive model $\model:\pastspace\to\probfuturespaceall$
    \caption{Attention CAPO}
    \label{algo:ourattentionmethod}
\end{algorithm}

\subsection{Counterfactual Action-Discrepancy CAPO}

Our second proposal can also be formulated as a re-weighted maximization objective, where we weight the log likelihood of each agent's trajectory in a scene by its \textit{individual} contribution to the ego's control decision. 
We do this by first enumerating through each agent in a scene, and computing counterfactual outputs from the AV's controller if every agent traversed their individual trajectory as recorded in the replay buffer, except for agent $n$. If we resample the trajectory that the $n$th agent \textit{might otherwise have taken}, $\predfuturek_n\sample\model(\Future_n|\past)$, we can compute the control output that would result:
\begin{equation}
\predactionk_{n} \;=\; \planner(\{\predfuturek_n\} \cup \future \setminus \{\futuren\}), \label{eq:whatif-upred}
\end{equation}
to compare against the control had no agent deviated from their recorded trajectories:
\begin{equation}
\action \;=\; \planner(\future). \label{eq:whatif-utrue}
\end{equation}
The difference in these two hypothetical controls corresponds to how much an individual agent affects the ego vehicle, and can represent the concern associated with predicting this particular agent in this particular instance accurately. If the model is probabilistic, then taking multiple samples $(K>1)$ helps ensure high importance even if the other agent only \textit{might} cause a control deviation: 
\begin{equation}
w_n \;=\; \max_{k\in\{1..K\}}||\action - \predactionk_n||_1, \label{eq:whatif-weight}
\end{equation}
which we use as weights for predictive model training:
\begin{equation}
\theta^* \;=\; \argmax_\theta 
\sum_{n=1}^N w_n \log \model(\futuren|\past). \label{eq:whatif-loss}
\end{equation}

We summarize our counterfactual action discrepancy method in \cref{algo:ourmethodwhatif}.
One benefit of this approach compared to the attention CAPO is that it is 
\begin{algorithm}[H]
    \textbf{Input}: Controller: $\planner:\pastspace\to\actionspace$
    \begin{algorithmic}[1]
     \STATE{Record trajectory data $\data=\{\past,\future\}_i$}
     \WHILE{training}
        \STATE{Sample batch $\past,\future\sample\data$}
        \STATE{Compute hypothetical controls: $\action,\predactionk_{n}$ \hfill $\triangleright$ \cref{eq:whatif-upred}--\eqref{eq:whatif-utrue}}
        \STATE{Compute weight: $w(\action,\predactionk_{n})$ \hfill $\triangleright$ \cref{eq:whatif-weight}}
        \STATE{Update model: $\theta\leftarrow\theta+ w(\action,\predactionk_{n})\nabla_{\theta}\log\model(\future|\past)$}
     \ENDWHILE
    \end{algorithmic}
    \textbf{Output}: Predictive model $\model:\pastspace\to\probfuturespace$
    \caption{Counterfactual CAPO}
    \label{algo:ourmethodwhatif}
\end{algorithm}

\subsection{Summary of Objectives}

There are various choices for utilities, or weights for traditional module metrics. In \cref{table:ourmethods} we summarize the several baselines methods, including NLL and our novel proposals.

\newcommand{\gainy}[0]{R2P2 $\text{Gain}_{\future}$}
\newcommand{\gainplanner}[0]{R2P2 $\text{Gain}_{\planner}$}
\newcommand{\gainplannerLone}[0]{R2P2 $\text{Gain}_{\planner1}$}

\newcommand{\weightgradpred}[0]{R2P2 $\text{Weight}_{\nabla\predfuture}$}
\newcommand{\weightgradtrue}[0]{R2P2 $\text{Weight}_{\nabla\future}$}
\newcommand{\weightplanner}[0]{R2P2 $\text{Weight}_\planner$}
\newcommand{\weightplannerk}[0]{R2P2 $\text{Weight}_{\planner k}$}
\newcommand{\weightattention}[0]{Attention}

\begin{table}[h]
\vspace{4pt}
\caption{Comparison of utilities and weighted objectives.}
\label{table:ourmethods}
\begin{center}
\def\arraystretch{1.5}
\resizebox{\linewidth}{!}{
\begin{tabular}{lccc} 
 \hline
 Method & Utility or Weight & Objective $\mathcal{L}(\theta)$ \\
 \hline
 \textit{Baselines:} \\
 \weightattention & $\delta(\future-\predfuture)$ & $ \model(\future|\past)+ \model(\futureego|\past)$ \\ 
  \gainy & $\delta(\future-\predfuture)$ & $ \model(\future|\past)$ \\ 
 \gainplannerLone & $||\planner(\future) - \planner(\predfuture)||_1$ & $\mathbb{E}_{\predfuture} \left[||\planner(\future) - \planner(\predfuture)||_1\right]$ \\ 
 \weightgradpred & $||\nabla_{\predfuture} \planner(\predfuture)||_1$ & $\mathbb{E}_{\predfuture} \left[||\nabla_{\predfuture} \planner(\predfuture)||_1\right]\model(\future|\past)$ \\
 \weightgradtrue & $||\nabla_{\future} \planner(\future)||_1$ & $||\nabla_{\future} \planner(\future)||_1\model(\future|\past)$ \\
 \hline
 \textit{Ours:} \\
 \weightplanner & $||\planner(\future)\!-\!\planner(\predfuture)||_1$ & $\mathbb{E}_{\predfuture} \left[||\planner(\future)\!-\!\planner(\predfuture)||_1\right] \model(\future|\past)$ \\
 \weightplannerk & $\max_k ||\planner(\future)\!-\!\planner(\predfuturek)||_1$ & \!\!$\max_k ||\planner(\future)\!-\!\planner(\predfuturek)||_1 \model(\future|\past)$ \\
 \weightattention Weight & $\alpha(\past)$ & $\alpha(\past)\model(\futureagent|\past) + \model(\futureego|\past)$ \\
 \hline
\end{tabular}
}
\end{center}
\end{table}
%
\section{Experimental Evaluation} \label{sec:experiments}

To evaluate our proposed method, we consider a representative scenario that is commonplace in autonomous driving: pedestrian trajectory prediction. The majority of pedestrian behaviors can safely be ignored by the AV’s autonomy stack; however, in rare cases of pedestrian-ego interaction (e.g., road crossings), accurate prediction of pedestrian behavior becomes crucial in avoiding collisions. This sparsity of interaction showcases how predictive models may perform well with respect to traditional metrics (e.g., ADE) while still leading to suboptimal ego behavior when it matters most.
Here, we first detail our experimental evaluation and implementation of the aforementioned scenario within the CARLA autonomous driving simulator \cite{dosovitskiy2017carla}. We then compare results between our method and the various baselines discussed in \cref{table:ourmethods}, where our experiments show that predictive models trained using our CAPO methods 
produce safe behavior with fewer collisions relative to other baselines.

\begin{figure}[h]
     \centering
     \begin{subfigure}[b]{0.49\linewidth}
         \centering
         \includegraphics[width=\linewidth]{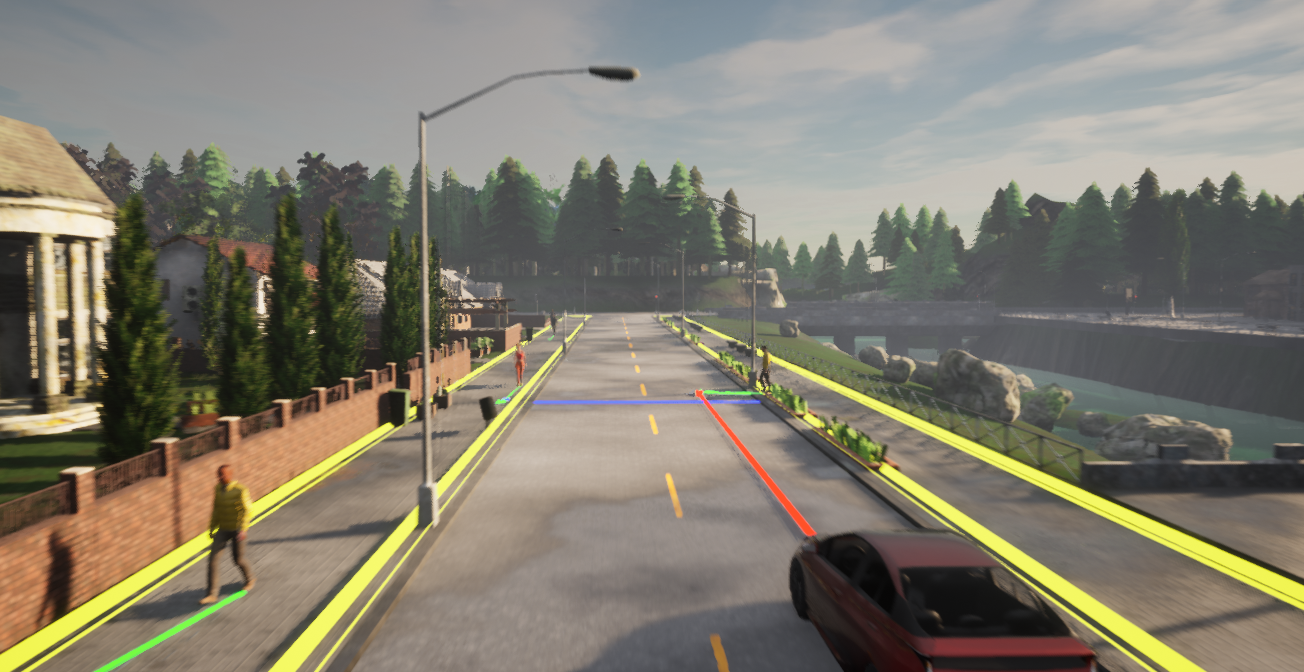}
     \end{subfigure}
      \hfill
     \begin{subfigure}[b]{0.49\linewidth}
         \centering
         \includegraphics[width=\linewidth]{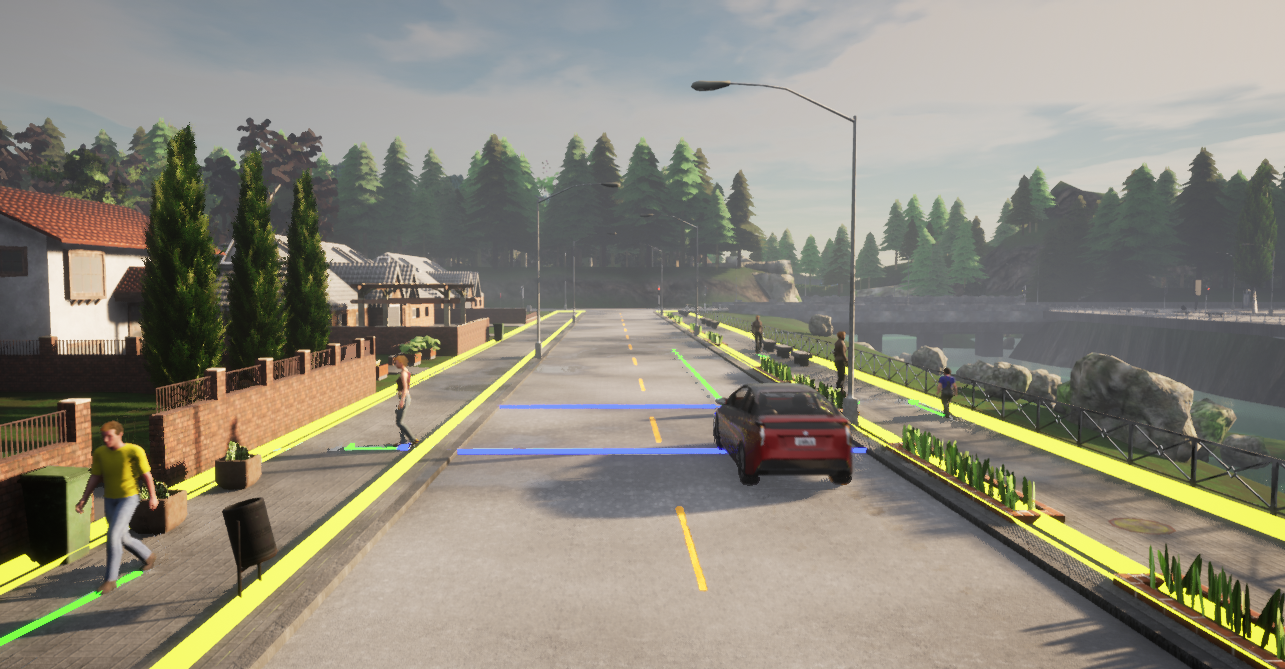}
     \end{subfigure}
        \caption{\textbf{Pedestrian Prediction Scenario}.
        Pedestrians spawn on the sidewalk (yellow region) and the ego (red car) predicts the pedestrian trajectories within the next 3 seconds (green). Some pedestrians will cross the road at right angles (blue). \textbf{Left}: the planner predicts a collision with a crossing pedestrian and starts slowing (red ego prediction up to blue line but not further). \textbf{Right}: ego is safely passing the road segment where the pedestrian has already crossed.
        }
        \label{fig:scenario1}
        \vspace{-6pt}
\end{figure}

\subsection{CARLA Scenario Design}
We implement our pedestrian prediction scenario in the CARLA simulator's Town01. A single ego vehicle is commanded to drive down a road that is adjacent to sidewalks which are populated with pedestrians. Occasionally, a pedestrian will cross the street and the ego agent must slow to avoid a collision when necessary.

The ego vehicle adjusts its longitudinal control to balance the competing priorities of maintaining the current speed limit (45mph) while avoiding collisions with the pedestrians crossing the road (either their current or future-predicted distance on the road ahead). This balance is performed by an Intelligent Driver Model \cite{treiber2000congested}. It uses predictions to estimate the closest collision distance and controls the vehicle to stop ahead of that point.

Pedestrians spawn at random locations on the sidewalk and are then provided a long-range navigation goal that is also uniformly sampled from the sidewalk. When the long-range goal is reached, another is sampled to replace it. To induce pseudo-random motion, a short-range goal is also generated at each time step. This goal is generated by projecting a point 4m along the path to the long-range goal, starting at the pedestrian’s location. The lateral offset $\beta_{t+1}$ of the short-range goal is generated by sampling from a normal distribution centered about the previous lateral offset $\beta_t$ after it has been scaled down (to drive it towards the long-range goal):
\begin{equation}
   \beta_{t+1} = (1-\varepsilon)\beta_t +  \mathcal{N}(0, \sigma^2),
\end{equation}
where $\sigma$ is the variance of the noise, and $\epsilon\in[0,1)$ is the commitment to the long-range goal.

When on the sidewalk, pedestrians are programmed to walk at speeds sampled about 2m/s while navigating around other pedestrians to avoid collisions and, occasionally, will pause outside of shops. Each different kind of pedestrian is defined with various noise levels, commitment, and stopping chance.
 
Pedestrians may also randomly decide to cross the road. The probability increases if their velocity vector points towards the road and increases greatly when the pedestrian is close to the road.
While crossing, they travel at 2 m/s in the shortest path possible, i.e., perpendicular to the road direction. To increase task difficulty, the probability that pedestrians will cross the road is increased at test time.

\begin{table*}[t]
\centering 
\vspace{4pt}
\caption{Scenario results, 100 episodes. Arrows indicate higher/lower preferred. Standard errors shown. \textbf{Best}, \underline{second}.}  
\label{tab:pedestrian-results}
\resizebox{\textwidth}{!}{
\begin{tabular}{lcccccc} 
 \hline
 Predictive Model & Success Rate $\uparrow$ & Collisions $\downarrow$ & Speed (m/s) $\uparrow$ & Jerk (m/$\text{s}^{-3}$) $\downarrow$ & ADE (m) $\downarrow$ & Control Error $\downarrow$ \\
 \hline
 \textit{Baselines:} \\
  \gainy & $89.0\%$ & $11$ & $9.97$ \scriptsize{$\pm 0.222$} & $8.92$ \scriptsize{$\pm 0.250$} & $2.09$ \scriptsize{$\pm 0.024$} & $0.59$ \scriptsize{$\pm 0.012$}\\ 
 \gainplannerLone & $85.0\%$ & $14$ & $10.45$ \scriptsize{$\pm 0.268$} & $6.65$ \scriptsize{$\pm 0.196$} & $3.48$ \scriptsize{$\pm 0.038$} & $0.63$ \scriptsize{$\pm 0.016$}\\ 
 \weightgradpred & $\underline{94.0}\%$ & $\underline{4}$ & $9.53$ \scriptsize{$\pm 0.216$} & $8.21$ \scriptsize{$\pm 0.140$} & $\mathbf{1.98}$ \scriptsize{$\pm 0.024$} & $0.60$ \scriptsize{$\pm 0.012$}\\ 
 \weightgradtrue & $91.0\%$ & $9$ & $9.74$ \scriptsize{$\pm 0.216$} & $8.74$ \scriptsize{$\pm 0.184$} & $\underline{2.00}$ \scriptsize{$\pm 0.025$} & $0.60$ \scriptsize{$\pm 0.011$}\\ 
\weightattention \, \cite{mercat2020multi} & $89.0\%$ & $11$ & $\underline{13.79}$ \scriptsize{$\pm 0.214$} & $\underline{4.48}$ \scriptsize{$\pm 0.147$} & $2.61$ \scriptsize{$\pm 0.050$} & $0.63$ \scriptsize{$\pm 0.026$}\\
 \hline
 \rowcolor{ForestGreen!30}  
 \textit{Oracle distribution} & $98.0\%$ & $2$ & $10.54$ \scriptsize{$\pm 0.231$} & $6.80$ \scriptsize{$\pm 0.180$} & $1.58$ \scriptsize{$\pm 0.036$} & $0.51$ \scriptsize{$\pm 0.013$}\\ 

\hline
\textit{Our methods:} \\
\weightplanner & $93.0\%$ & $7$ & $8.86$ \scriptsize{$\pm 0.188$} & $9.26$ \scriptsize{$\pm 0.194$} & $2.29$ \scriptsize{$\pm 0.022$} & $\underline{0.58}$ \scriptsize{$\pm 0.010$}\\ 
 \weightplannerk & $\mathbf{99.0}\%$ & $\mathbf{1}$ & $9.46$ \scriptsize{$\pm 0.196$} & $7.89$ \scriptsize{$\pm 0.159$} & $2.14$ \scriptsize{$\pm 0.018$} & $\mathbf{0.55}$ \scriptsize{$\pm 0.011$}\\ 
 
\weightattention \, Weight$_\attention$  & $91.0\%$ & $9$ & $\mathbf{14.36}$ \scriptsize{$\pm 0.217$} & $\mathbf{4.22}$ \scriptsize{$\pm 0.154$} & $2.58$ \scriptsize{$\pm 0.053$} & $0.64$ \scriptsize{$\pm 0.024$}\\  
  \hline
\end{tabular}
}
\vspace{-11pt} 
\end{table*}

\subsection{Compared models}

   1) \textbf{Oracle distribution} The pedestrian behavior is simulated with a known distribution at each time step, which is sampled to produce a trajectory. The trajectory distribution is approximated by sampling $K=5$ trajectories for each pedestrian. The planner reacts to the trajectory that would cause the closest intersection with its desired path. This method is a perfect probabilistic predictor which is only accessible with simulated data.
   Its predictions are not biased towards sampling the most critical trajectories, and planning with relatively few samples can yield suboptimal results.

  2) \textbf{Gradient weighting} Recent work has also investigated weighing prediction objectives by a measure of local sensitivity of downstream costs to individual predictions \cite{ivanovic2021rethinking}. This method first learns a cost function to evaluate ego controls given predicted human trajectories, using inverse optimal control. The method then weights each prediction loss by the gradient magnitude of the cost outputs w.r.t.\ the prediction inputs. In our work, we consider using the controller directly, forming an analogous baseline: using gradients of the controller w.r.t.\ the predicted trajectory $||\nabla_{\predfuture} \planner(\predfuture)||_1$, or true trajectory $||\nabla_{\future} \planner(\future)||_1$, included in \cref{table:ourmethods}. While we do not assume differentiable controllers in our own methods, we nevertheless experiment using a differentiable controller to compare against this baseline method. 
    \newpage
    
   3) \textbf{Attention weighting} As presented in section \ref{sec:attention}. We train this model with our CAPO method (algorithm \ref{algo:ourattentionmethod}) and, as a baseline, we compare it with the unbiased prediction as the predictor \cite{mercat2020multi} would produce.

    4) \textbf{Reparameterized Pushforward Policy (R2P2)} we use the likelihood-based multi-agent prediction algorithm R2P2 \cite{rhinehart2018r2p2} as baseline $\text{Gain}_{\future}$, and also use R2P2 as the base model for all other predictive models apart from the attention model. R2P2 is a autoregressive normalizing flow, capable of expressing multimodal agent trajectories, trained with NLL. We parameterize R2P2 to predict 30 steps with data at 10Hz, corresponding to a 3s prediction for all pedestrians. We train it with our CAPO method (algorithm \ref{algo:ourmethodwhatif}) using $K=10$ samples and we use $K=1$ samples at test time.

\subsection{Metrics}

The table \ref{tab:pedestrian-results} presents our results for 100 sequences. We track the performance of the system (prediction and planner) with the success rate and the number of collisions.
Three conditions may end a sequence: 
\begin{itemize}
    \item Success: vehicles traverses 200m road without incident.
    \item Collision: a pedestrian was hurt.
    \item Time out: the car was too slow ($>60$s).
\end{itemize}

We also score efficiency and comfort indicators by average speed and average jerk respectively.
Finally, we compute the average pedestrian trajectory prediction errors and their downstream effect on the planner with the average displacement error (ADE) and the \textit{Control Error} equal to $||\planner(\future) - \planner(\predfuture)||_1$.
The control error measures the downstream effect of the prediction error on the ego's plans.

\section{Discussion} 

The results in \cref{tab:pedestrian-results} show that while all methods do reasonably well, weighting predictive objectives by their downstream effect improves the downstream performance as illustrated by a low collision count and control error. While methods such as \weightgradpred\ assume a differentiable controller, we find this assumption does not need to be made, and our methods can work with any type of controller.
While our methods did not score as well on the ADE metric of agents' trajectories, they did score best on the metrics that matters more: the control error, and success rate, thus mitigating error propagated downstream and improving the end task performance.
We compared to objectives weighted by the planner's sensitivity  $||\nabla_{\predfuture} \planner(\predfuture)||_1$, which is related to~\cite{ivanovic2021rethinking}. However, such methods assume the planner (or cost function) is differentiable, which is often not the case in real AV systems. Secondly, \textit{local} sensitivity to a point prediction is not necessarily a good measure of relevance if the gradient is noisy or changes drastically over the full predictive distribution.
While our experiments show encouraging results, testing with various setups and environments would be needed to give a clear best method.
\section{Conclusions} \label{sec:conclusions}
Modular autonomous systems (such as those commonly used in AVs) provide a number of advantages, but generally incur the disadvantage that individual components typically do not optimize for system-wide or downstream performance metrics directly.
In this paper, we proposed weighted objectives for learning prediction models that account for the downstream objective without imposing stringent requirements on downstream components (such as end-to-end differentiability).
These objectives weight the usual likelihood objective, either using attention weights derived from a behavior-cloned policy, or using the impact that substituting predicted trajectories for ground-truth trajectories has on planner output.
Accounting for the downstream objective in this manner encourages prediction models to focus on what's important -- either at the agent or individual trajectory level -- and, as a result, improves system-wide performance, as we showed empirically in a pedestrian jaywalking scenario.

A number of promising avenues exist for future research. 
First, control-aware objectives may provide out-of-domain generalization benefits by encouraging prediction models to focus on relevant aspects of the scene, and ignore spurious sources of information that are safe to ignore.
Second, in this paper we focused on data collected from an expert. However, this requirement limits the applicability of the proposed metrics, and a broader coverage of the state-space 
resulting from the use of both expert and suboptimal data might improve the learned prediction models.
Finally, although we focused in this paper on using control-aware weighting for \textit{optimizing} prediction models. Our method might equally well be used to define a weighted metric for \textit{evaluating} models in a validation setting where training-time access to the downstream planner is either not available or undesirable.

\newpage

\printbibliography

\end{document}